\documentclass[conference]{IEEEtran}

\IEEEoverridecommandlockouts

\makeatletter
 \let\old@ps@headings\ps@headings
 \let\old@ps@IEEEtitlepagestyle\ps@IEEEtitlepagestyle
 \def\confheader#1{%

 \def\ps@IEEEtitlepagestyle{%
 \old@ps@IEEEtitlepagestyle%
 \def\@oddhead{\strut\hfill#1\hfill\strut}%
 \def\@evenhead{\strut\hfill#1\hfill\strut}%
 }%
 \ps@headings%
 }
 \makeatother

 \usepackage[pscoord]{eso-pic}

\usepackage{cite}
\usepackage{amsmath,amssymb,amsfonts}
\usepackage{algorithmic}
\usepackage{graphicx}
\usepackage{textcomp}
\usepackage{xcolor}
\usepackage{diagbox}
\usepackage{eso-pic}
\makeatletter
\newcommand\subparagraph{\@startsection{subparagraph}{5}{\parindent}%
  {1ex \@plus1ex \@minus .2ex}%
  {-1em}%
  {\normalfont\normalsize}}
\makeatother
\usepackage{multirow}
\usepackage{lipsum}
\usepackage[utf8]{inputenc}
\usepackage{tikz}
\usetikzlibrary{shapes,arrows}
\usetikzlibrary{positioning}
\usepackage{subcaption}
\usepackage{tabularx}
\usepackage{subcaption}
\usepackage{fontspec}
\usepackage{polyglossia}
\newfontfamily\bengalifont[Script=Bengali]{kalpurush.ttf}
\usepackage{caption}
\usepackage{enumitem}
\usepackage{array}
\usepackage[hidelinks]{hyperref}

\DeclareCaptionLabelFormat{figlabel}{\textnormal{Fig. #2}}
\DeclareCaptionFormat{figformat}{\textnormal{#1#2#3}}
\def\BibTeX{{\rm B\kern-.05em{\sc i\kern-.025em b}\kern-.08em
    T\kern-.1667em\lower.7ex\hbox{E}\kern-.125emX}}

\begin{document}

\AddToShipoutPictureBG*{%
  \AtPageLowerLeft{%
    \raisebox{0.5cm}{\makebox[\paperwidth]{%
      \parbox[b]{0.85\paperwidth}{\scriptsize\centering
        \copyright~2023 IEEE. Personal use of this material is permitted. Permission from IEEE must be obtained for all other uses, in any current or future media, including reprinting/republishing this material for advertising or promotional purposes, creating new collective works, for resale or redistribution to servers or lists, or reuse of any copyrighted component of this work in other works.\\
        Accepted version of the paper published in the 2023 6th International Conference on Electrical Information and Communication Technology (EICT), DOI: 10.1109/EICT61409.2023.10427730.}}}}}

\title{Color Independent Word Segmentation From Transcribed Bangla Passages} 

\author{\IEEEauthorblockN{Faias Satter}
\IEEEauthorblockA{\textit{Department of Computer Science and Engineering} \\
\textit{Khulna University of Engineering \& Technology }\\
 Khulna-9203, Bangladesh\\
faiaspromit@gmail.com}
\and
\IEEEauthorblockN{Noor Masrur}
\IEEEauthorblockA{\textit{Department of Computer Science and Engineering} \\
\textit{Khulna University of Engineering \& Technology}\\
Khulna-9203, Bangladesh \\
noor.masrur.bd@gmail.com}
\and
\IEEEauthorblockN{Sk. Md. Masudul Ahsan}
\IEEEauthorblockA{\textit{Department of Computer Science and Engineering} \\
\textit{Khulna University of Engineering \& Technology}\\
Khulna-9203, Bangladesh \\
smahsan@cse.kuet.ac.bd}

}
\author{\IEEEauthorblockN{Faias Satter\IEEEauthorrefmark{1}, Noor Masrur\IEEEauthorrefmark{2}, Sk. Md. Masudul Ahsan\IEEEauthorrefmark{3}}
\IEEEauthorblockA{Department of Computer Science and Engineering\\
Khulna University of Engineering \& Technology\\
Khulna-9203, Bangladesh\\
Email: \IEEEauthorrefmark{1}faiaspromit@gmail.com \IEEEauthorrefmark{2}noor.masrur.bd@gmail.com \IEEEauthorrefmark{3}smahsan@cse.kuet.ac.bd}
}

\maketitle

\begin{abstract}
An optical character recognition(OCR) system can scan paper and extract text, making people's jobs easier. While numerous OCR systems are accessible in the software sector, finding a dependable equivalent solution for Bangla is tough. When it comes to handwritten texts, the case is even more rare. The first fundamental step to any OCR is to segment words from text images. If this stage fails, the total OCR's performance will be poor no matter how promising the later stages perform. This research aims to segment words in a handwritten Bangla text image. This research can be implemented on any smartphone-captured image, irrespective of the color and type of paper and ink. Furthermore, as smartphone-captured images can create shadow interferences, the custom dataset built for this research is created in such a way that every possible obstacle that can be faced is included. For 7374 words, a total of 7278 bounding boxes are generated, which have recall of 90.60 \%, precision of 91.80 \%, and F1-score of 91.20 \%. The system can be further improved with nested operations on bounding boxes containing several words or by adjusting the adaptive thresholding and dilation filter sizes to a more precise level.
\end{abstract}


\section{Introduction}
One of the computer vision applications is optical character recognition (OCR), which converts images of typed/handwritten/printed text into machine-encoded text. OCR is commonly used to digitize texts in an image so that they can be electronically edited, searched, stored more compactly, and used in machine processes like cognitive computing, machine translation, and text-to-speech. However, there is currently no reliable OCR for the Bangla language, the eighth most spoken language in the world \cite{id19}. The first step for any OCR system is to segment words from texts. Therefore, the authors aim to develop a word segmentation system specifically for detecting handwritten Bangla words from images of paper documents.

The authors aim to scan an image of handwritten Bangla texts, irrespective of the paper/ink color and shadow interferences, detect the words, and segment them using different image processing techniques. However, handwritten Bangla characters are not as straightforward as English characters. Sometimes, the handwritten letters from different words are placed so closely that it is difficult to distinguish them as separate words. Moreover, characters in Bangla are connected by a line within a word called Matra, so they do not have separate spaces inside a handwritten Bangla word like in English. Nevertheless, some characters do not have Matra. So, the gaps can be misread as separate words. Furthermore, while dealing with Bangla vowels, e.g., \textbf{\bengalifont'ু', 'ূ', 'ি', 'ী', 'ৈ', 'ৌ'} pieces of one word can be absorbed by another. These limitations make developing an image processing system for Bangla handwriting more complicated.

This system aims to present the authors' efforts to use image processing techniques to develop a bespoke dataset and optimize the performance of word detection from handwritten Bangla paragraphs. Although there is yet to be a high-quality OCR for Bangla like there is for English, Chinese, or Arabic, the authors hope their system will be a milestone toward meeting the demand for reliable OCR in Bangla. The scope of this paper encompasses the segmentation of words in handwritten Bangla text images.

\section{Related Works}
Bishnu et al.\cite{id1} used a recursive contour following method to determine character extents in one zone. Pal et al.\cite{id2} proposed a water reservoir-based approach for numeral segmentation. Sharda et al.\cite{id3} implemented skew detection correction using the scan line and Hough transform methods. Sekehravani et al.\cite{id4} applied the Canny edge detection algorithm with a median filter to noisy images. Duan et al.\cite{id5} proposed an improved Hough transform for line detection. Lu et al.\cite{id6} reviewed character extraction strategies for handwritten English word recognition. Lee et al.\cite{id7} used gray-scale image properties and a multi-stage graph search algorithm for character segmentation and recognition. Dave\cite{id8} compared different segmentation methods for handwritten character identification and found the histogram approach to be the most effective, but it does not work with handwritten text images. Zhou et al.\cite{id9} developed a text identification system using color clustering and linked component analysis. Sulem et al.\cite{id10} proposed a method for detecting text lines on handwritten sheets using Hough Transformation.

Sarkar et al.\cite{id12} presented a novel approach for word extraction in unconstrained handwritten Bangla document images of text lines. However, their work was limited to binary images as they worked on the CMATERdb1.1.1 Dataset\cite{id17} only, and their success rate was 86.01\%. Mullick et al.\cite{id13} proposed a method for effectively segmenting lines of handwritten Bangla text in document images. However, as this method initially depends on estimating the number of lines, this system failed to segment if there is a significant difference between the extents of two lines. Rakshit et al.\cite{id14} aimed to develop a segmentation method for separately extracting lines, words, and characters from handwritten Bangla text. However, their average result was only 81.47\%. Das et al.\cite{id15} proposed a method combining horizontal projection profile, seam carving, and contour tracing for line and word segmentation of language-independent handwritten documents. However, their average accuracy was only 84.55\%. This result degraded more if text lines have converse skew angles. Bhattacharya et al.\cite{id16} presented an end-to-end system for Bangla online handwriting recognition. The line segmentation fails if part of two lines falls into the same x-axis. Hasan et al.\cite{id18} proposed an image segmentation-based word segmentation method in their work with an accuracy of 89.45 \%. However, they tested against only 50 images of handwritten texts, which generated only 3655 bounding boxes and worked on only binary images without shadow interferences.

\section{Proposed Methodology}


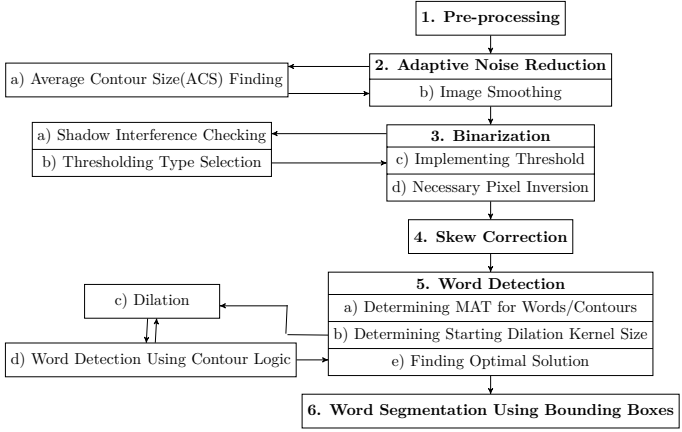
\begin{figure}
    \centering
    \resizebox{0.5\textwidth}{!}{
    \begin{tikzpicture}[node distance=0.4cm, every node/.style={rectangle, draw, minimum width=3cm, minimum height=0.75cm},
    double/.style={draw, anchor=text, rectangle split,rectangle split parts=2},
    triple/.style={draw, anchor=text, rectangle split,rectangle split parts=3},
    quadruple/.style={draw, anchor=text, rectangle split,rectangle split parts=4},
    pentuple/.style={draw, anchor=text, rectangle split,rectangle split parts=5},
    hextuple/.style={draw, anchor=text, rectangle split,rectangle split parts=6},>=stealth']

      \node (step1) {\textbf{1. Pre-processing}};
      \node (step2) [double,below=of step1] {\textbf{2. Adaptive Noise Reduction}
        \nodepart{second}
        b) Image Smoothing
      };
      \node (leftofstep2) [left=of step2, yshift=0cm, xshift = -1.4cm] {a) Average Contour Size(ACS) Finding}; 
      \node (step3) [triple,below=of step2] {\textbf{3. Binarization}
        
        \nodepart{second}
        c) Implementing Threshold
        \nodepart{third}
        d) Necessary Pixel Inversion
      };
      \node (leftofstep3) [double,left=of step3, yshift=0.3cm, xshift = -2.15cm] {
        a) Shadow Interference Checking
        \nodepart{second}
        b) Thresholding Type Selection}; 
      
      \node (step4) [below=of step3] {\textbf{4. Skew Correction}};
      \node (step5) [quadruple,below=of step4] {\textbf{5. Word Detection }
        \nodepart{second}
        a) Determining MAT for Words/Contours
        \nodepart{third}
        b) Determining Starting Dilation Kernel Size
        \nodepart{fourth}
        e) Finding Optimal Solution
      };
      \node (leftofstep5--1) [left=of step5, yshift=0.5cm, xshift = -2cm] {c) Dilation};
      \node (leftofstep5--2) [left=of step5, yshift=-.8cm, xshift = -.3cm] {d) Word Detection Using Contour Logic}; 
      \node (step6) [below=of step5] {\textbf{6. Word Segmentation Using Bounding Boxes}};
    
      \draw [->] (step1) -- (step2);
      \draw [->] (step2) -- (step3);
      \draw [->] (step3) -- (step4);
      \draw [->] (step4) -- (step5);
      \draw [->] (step5) -- (step6);
      \draw [->] ([xshift=0cm,yshift=0.3cm]step2.west) -- ([xshift=0cm,yshift=0.3cm]leftofstep2.east);
      \draw [->] ([xshift=0cm,yshift=-0.3cm]leftofstep2.east) -- ([xshift=0cm,yshift=-0.3cm]step2.west);
      \draw [->] ([xshift=0cm,yshift=0.65cm]step3.west) -- ([xshift=0cm,yshift=0.35cm]leftofstep3.east);
      \draw [->] ([xshift=0cm,yshift=-0.3cm]leftofstep3.east) -- ([xshift=0cm,yshift=-0cm]step3.west);
      \draw [->] ([xshift=-0.9cm,yshift=0.4cm]step5.west) -- ([xshift=0cm,yshift=-0.1cm]leftofstep5--1.east);
      \draw [-] ([xshift=0cm,yshift=-.25cm]step5.west) -- ([xshift=1.5cm,yshift=-0.74cm]leftofstep5--1.east);
      \draw [-] ([xshift=3cm,yshift=.3cm]leftofstep5--1.south) -- ([xshift=3cm,yshift=0.2cm]leftofstep5--2.north);
      \draw [->] ([xshift=0cm,yshift=0cm]leftofstep5--2.east) -- ([xshift=0cm,yshift=-.8cm]step5.west);
      \draw [->] ([xshift=-0.1cm,yshift=0cm]leftofstep5--1.south) -- ([xshift=-0.1cm,yshift=0cm]leftofstep5--2.north);
      \draw [->] ([xshift=0.1cm,yshift=0cm]leftofstep5--2.north) -- ([xshift=0.1cm,yshift=0cm]leftofstep5--1.south);

    \end{tikzpicture}
    }
  \caption{Flowchart of the working procedure}
  \label{fig1}
\end{figure}

\subsection{Working Procedure}
The total working procedure of the system can be divided into some consequent parts. The working procedure follows the flowchart in Fig. \ref{fig1}. Not every section is directly implemented to gain the output; some sections act as a requisite part of other sections. The requisite sections are drawn on the left side of the flowchart, and the fundamental parts are drawn on the right side. 
\subsubsection{Pre-processing}
\begin{figure}
  \centering
  \resizebox{0.5\textwidth}{!}{
      \begin{subfigure}[b]{0.49\textwidth}
        \centering
        \includegraphics[width=\textwidth,height=120pt]{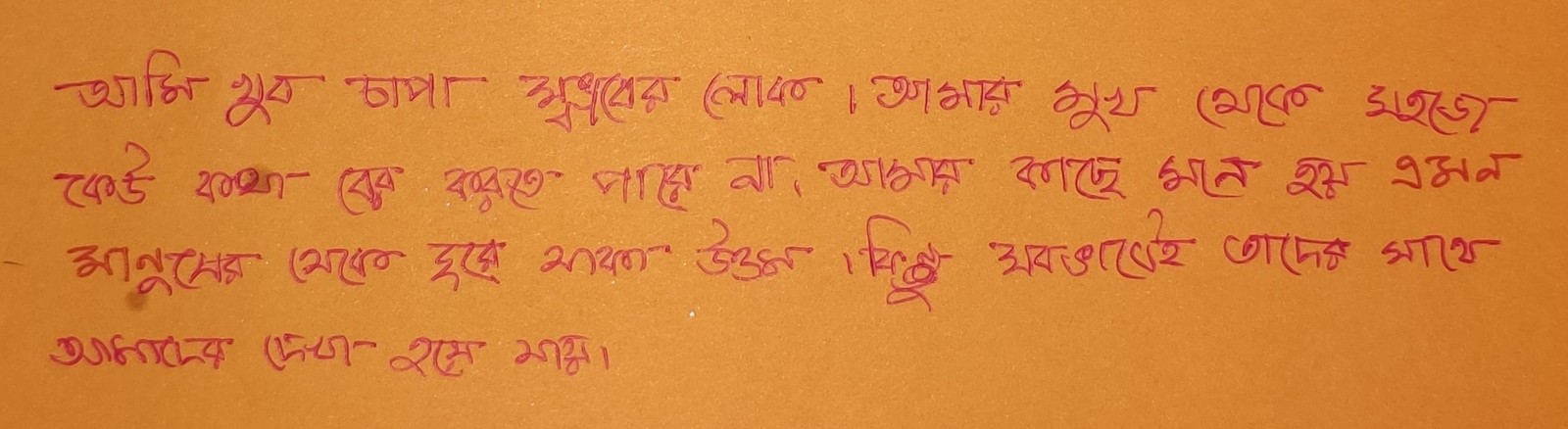}
        \captionsetup{font=Huge}
        \caption{\huge Input Image}
        \label{fig2a}
      \end{subfigure}
      \hfill
      \begin{subfigure}[b]{0.49\textwidth}
        \centering
        \includegraphics[width=\textwidth,height=120pt]{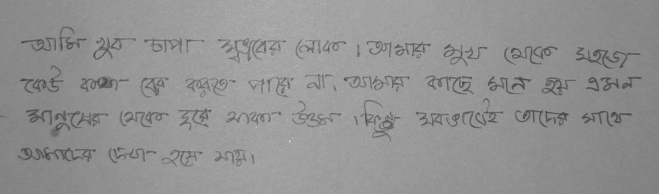}
        \captionsetup{font=Huge}
        \caption{\huge Grayscale Image}
        \label{fig2b}
      \end{subfigure}
      \hfill
      \begin{subfigure}[b]{0.49\textwidth}
        \centering
        \includegraphics[width=\textwidth,height=120pt]{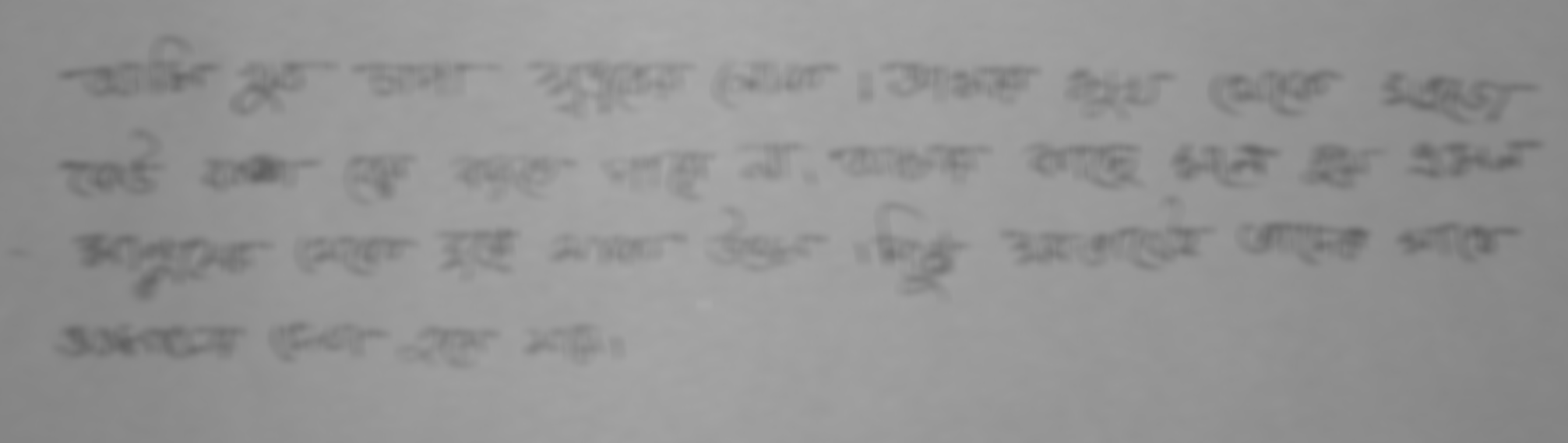}
        \captionsetup{font=Huge}
        \caption{\huge Smoothed Image }
        \label{fig2c}
      \end{subfigure}
    } 
  \caption{Input image to smoothed Grayscale image}
  \label{fig2}
\end{figure}

The image is scanned in grayscale mode to achieve better results. Although scanning in black and white allows for faster scanning, some character information may be lost. Fig. \ref{fig2a} and \ref{fig2b} demonstrates the input and output images after grayscale transformation.\\

\subsubsection{Adaptive Noise Reduction}
Different image-capturing devices,e.g., Smart Phones, capture images in different resolutions. The image sizes vary from device to device. Also, different people have different handwriting sizes. So, for noise reduction, different kernel sizes should be determined first.
\paragraph{Avergage Contour Size(ACS) Finding}The ACS of the words is determined first; for determining different kernel sizes for different images. The grayscale image is denoised first; then, adaptive thresholding is applied. In this stage, denoising uses a Gaussian kernel with a kernel size of 21 * 21, a standard deviation in the y-direction of 21, and a scaling factor on the standard deviation in the y-direction of 7. Adaptive thresholding takes an average value using a kernel size of 21 * 21. After thresholding, the words can be considered contours, and any contour with an area less than 50 is ignored as salt noise.
\paragraph{Image Smoothing}
The Non-local Means Denoising algorithm removes noise from images by replacing a pixel's value with an average of the values of several surrounding pixels. The equation of the kernel used here is, 
\begin{equation}
k[i, j] = \frac{{x^2 - \sigma^2 + y^2 - (\sigma \cdot \theta)^2}}{{2 \pi \sigma^5 (\sigma \cdot \theta)}} \cdot e^{-\frac{{x^2}}{{2\sigma}} - \frac{{y^2}}{{2(\sigma \cdot \theta)}}}
\end{equation}
where x and y represent the coordinates of each element in the kernel, Sigma is kernel size's square root and kernel size is obtained from ACS by an empirically decided mapping function (TABLE \ref{table1}). The table can be extended further, maintaining a continuous linear relationship(for every 200 increments in ACS, the kernel size is incremented by 2).
However, when white/less colorful paper(background pixel value between 181-255) is used, it generates less salt noise. But, as the background/paper color gets darker, the amount of salt noise increases. That is why a larger kernel size should be used. Therefore, additional action is taken on the table's result,e.g., if the background color gets darker,e.g., 101 - 180, the kernel size is further increased by two, and for less than 100, kernel size is further increased by 4. Fig. \ref{fig2c} is the denoised image of the grayscale image shown in Fig. \ref{fig2b}.      
\begin{table}[t]
  \centering
  \caption{Mapping of Average Contour Size to Kernel Size}
  \begin{tabular}{|c|c|}
    \hline
    \centering Average Contour Size (ACS) & \centering Kernel Size \tabularnewline
    \hline
    \centering ACS < 200 & \centering 3 * 3 \tabularnewline
    \hline
    \centering $200 \leq ACS < 300$ & \centering 5 * 5 \tabularnewline
    \hline
    \centering $300 \leq ACS < 400$ & \centering 7 * 7 \tabularnewline
    \hline
    \centering $400 \leq ACS < 500$ & \centering 9 * 9 \tabularnewline
    \hline
    \centering $500 \leq ACS < 600$ & \centering 11 * 11 \tabularnewline
    \hline
    \centering $600 \leq ACS < 700$ & \centering 13 * 13 \tabularnewline
    \hline
    \centering $700 \leq ACS < 800$ & \centering 15 * 15 \tabularnewline
    \hline
    \centering $800 \leq ACS < 900$ & \centering 17 * 17 \tabularnewline
    \hline
    \centering $900 \leq ACS < 1000$ & \centering 19 * 19 \tabularnewline
    \hline
    \centering $1000 \leq ACS < 1200$ & \centering 21 * 21 \tabularnewline
    \hline
    \centering $1200 \leq ACS < 1400$ & \centering 23 * 23 \tabularnewline
    \hline
  \end{tabular}
  
  \label{table1}
\end{table} 
\subsubsection{Binarization}
\begin{figure}
  \centering
  \resizebox{0.5\textwidth}{!}{
      \begin{subfigure}[b]{0.45\textwidth}
        \centering
        \includegraphics[width=\textwidth,height=75pt]{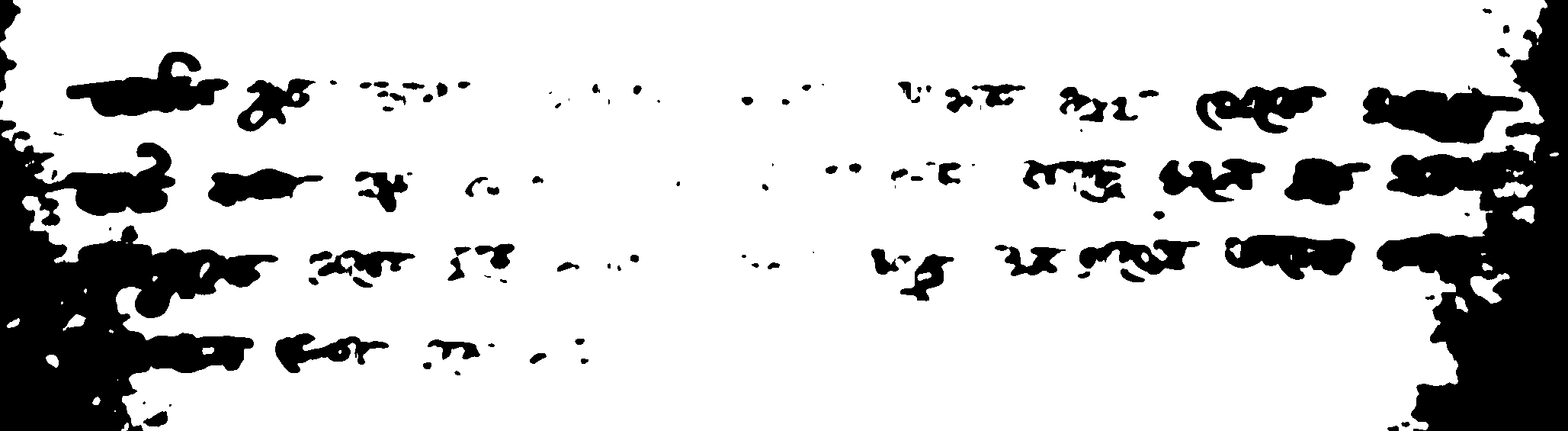}
        \captionsetup{font=LARGE}
        \caption{Otsu Thresholding}
        \label{fig4a}
      \end{subfigure}
      \begin{subfigure}[b]{0.45\textwidth}
        \centering
        \includegraphics[width=\textwidth,height=75pt]{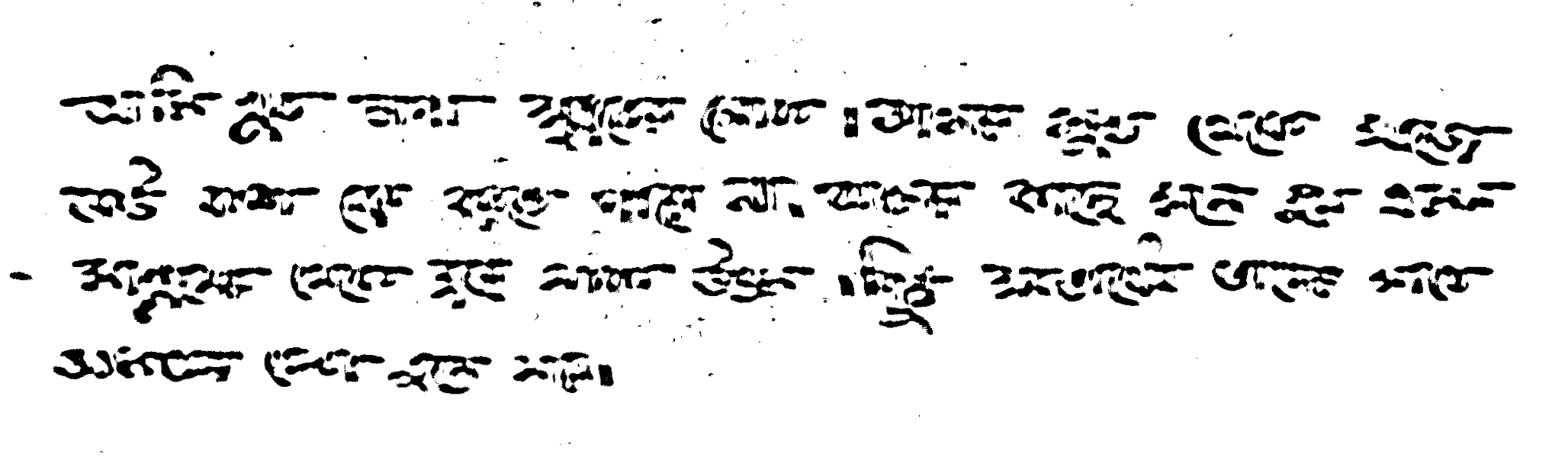}
        \captionsetup{font=LARGE}
        \caption{Adaptive Thresholding}
        \label{fig4b}
      \end{subfigure}
  }
  \caption{Result after applying both thresholding on Fig. \ref{fig2c}}
  \label{fig4}
\end{figure}

Thresholding divides pixel values into two groups, white for the foreground and black for the foreground. Either the Global Optimal Thresholding(Otsu) or Adaptive Thresholding method is used based on the characteristics of the image. 
\paragraph{Shadow Interference Checking} Before deciding what thresholding method to use, shadow interference is checked on the image. The area of the largest contour is obtained after applying Otsu on the image to determine the illumination/shadow issue in an image. If $\left(\frac{{\text{{Area of largest contour}}}}{{\text{{Average Contour Size}}}}\right) > 100$ then that means there is shadow in the image.
\paragraph{Thresholding Type Selection} Otsu or Adaptive thresholding is selected based on the conditions shown in (TABLE \ref{table8}).
\begin{table}[b]
  \centering
  \caption{Selection of Thresholding Type}
  \begin{tabularx}{0.4954\textwidth}{|m{1.8cm}|m{2cm}|m{1.7cm}|m{1.705cm}|}
    \hline
    \centering Background/ Page Color & \centering Colorful (pixels value $<$ 180) & \multicolumn{2}{|c|}{White(pixels value $\geq$ 180)} 
    \tabularnewline
    \hline
    \centering Shadow Interferences & \centering  & \centering Yes & \centering No 
    \tabularnewline
    \hline
    \centering Thresholding Type & \centering Adaptive & \centering Adaptive & \centering Otsu 
    \tabularnewline
    \hline
  \end{tabularx}
  
  \label{table8}
\end{table}
\paragraph{Implementing Threshold}
Fig. \ref{fig2a} has an illumination issue in the image, and the background pixel value = 133. Its output is shown in Fig. \ref{fig4} after applying both Otsu and Adaptive thresholding. The kernel size/window size for Adaptive thresholding is determined using the empirically decided mapping function shown in TABLE \ref{table1}. Although Adaptive thresholding applies to every kind of image, Otsu works best in the later stages, where one single threshold is enough for the whole image. Besides, Adaptive thresholding creates fragmented texts as it obtains threshold from a particular sized window while Otsu leaves the texts unaltered. In the case of fragmented texts, sometimes, part of the characters from a word can be excluded from the contour. 
\paragraph{Necessary Pixel Inversion} After thresholding, the pixel values are inverted where needed, creating a uniform scenario. Fig. \ref{fig7b} shows the result after the inversion applied on Fig. \ref{fig2c}. 
\subsubsection{Skew Correction}
This section involves identifying and correcting document skew, essential for accurate OCR. At first, the Canny edge detection algorithm \cite{id4} is applied. Then, the Hough transform method\cite{id5} is applied to identify co-linear lines as intersecting points, which are used to correct skew.

\subsubsection{Word Detection}
At first, dilation is applied using a squared structured element with white foreground pixels.

\paragraph{Determining Minimum Area Threshold(MAT) for Words/Contours} A rectangular bounding box detects contours/words as characters in Bangla are closely connected. Connected pixels with an area larger than a minimum threshold are considered a contour, and a bounding box is applied to wrap the contour. The dilation process works based on this minimum area threshold(MAT). To determine the MAT from ACS, two different mapping techniques are used based on the thresholding method applied previously. 
\subparagraph{Determining MAT after OTSU Thresholding}
In the case of Otsu thresholding, which does not create any fragmented characters, the MAT is calculated as follows: 
\begin{equation}
\text{Minimum Area Threshold} = \frac{\text{Average Contour Size}}{1.25}
\end{equation}
\subparagraph{Determining MAT after Adaptive Thresholding}
Since the letters are fragmented in the case of Adaptive thresholding, the ACS is much smaller than the actual average contour area. Besides, a new term called writing density(WD) is introduced. WD represents how densely words are distributed on a page and characterizes a person's handwriting style. The more the value of WD, the less dense the words are. 
\begin{equation}
\text{Writing Density} = \frac{\text{Total area of the page}}{\text{Average Contour Size}} 
\end{equation} 
Therefore, an empirically decided mapping function (TABLE \ref{table3}) maps the WD to the MAT. Furthermore, an additional empirically decided mapping function (TABLE \ref{table4}) is applied to the result of the previous one. 
\paragraph{Determining Starting Dilation Kernel Size}
A minimum area for words is determined first, and then dilation is applied. The challenge is in determining the starting kernel size for dilation; if it is too tiny, an under-segmentation error will be generated (Fig. \ref{fig10c}), and if it is too large, an over-segmentation error (Fig. \ref{fig10d}) will be generated. However, in case of an over-segmentation error, if two or more words merge, the total contour number will decrease. The optimal solution is when the total contour number is maximum. So, only considering the fact that when starting kernel is not too small is enough. TABLE \ref{table2} shows the empirically established mapping of ACS to starting kernel size of dilation for both the case of Otsu and Adaptive thresholding. This table continues further while maintaining a linear relationship(for every 2000 increments in ACS, the kernel size is incremented by 1).

\paragraph{Dilation} The dilation process is executed using the starting dilation kernel size and one iteration. 

\begin{figure}
  \centering
  \resizebox{0.5\textwidth}{!}{
      \begin{subfigure}[b]{0.5\textwidth}
        \centering
        \includegraphics[width=\textwidth, height = 70pt]{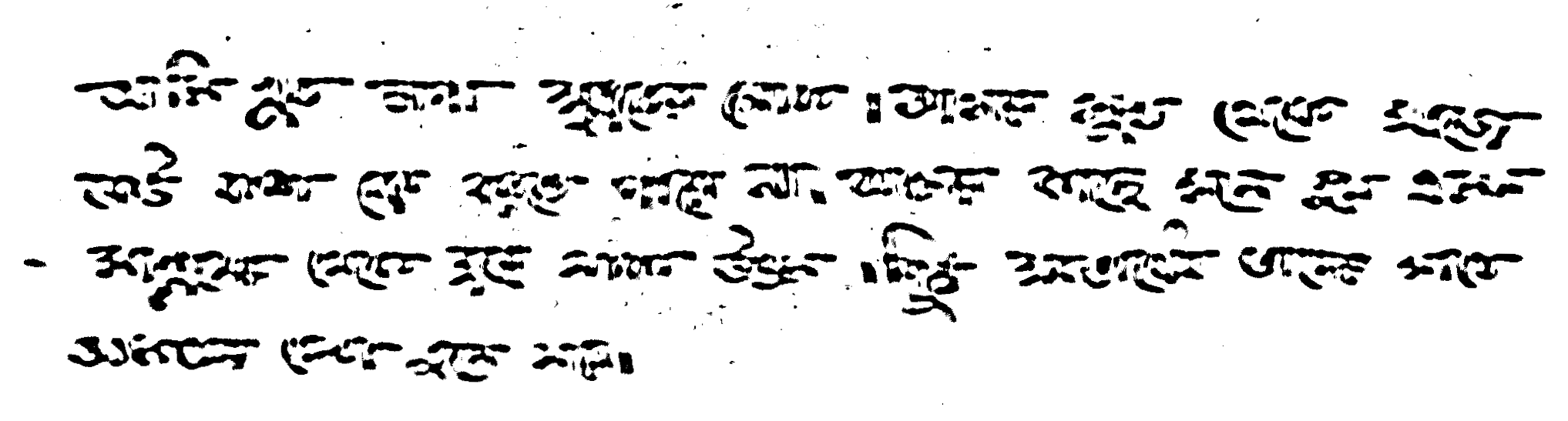}
        \captionsetup{font=huge}
        \caption{Thresholding on Fig. \ref{fig2a}}
        \label{fig7a}
      \end{subfigure}
      \hfill
      \begin{subfigure}[b]{0.5\textwidth}
        \centering
        \includegraphics[width=\textwidth, height = 70pt]{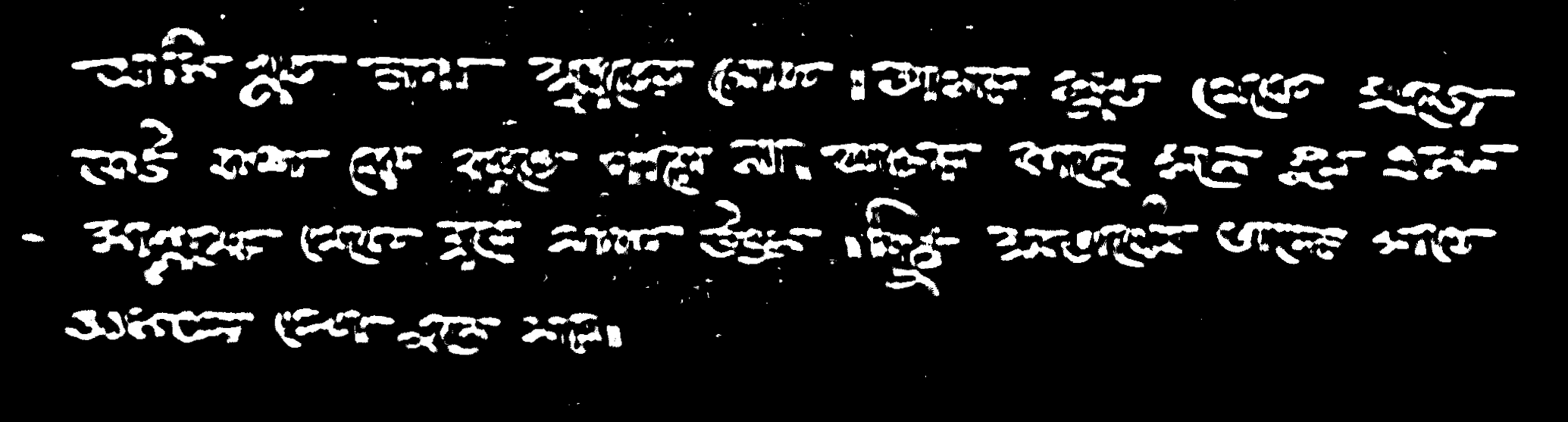}
        \captionsetup{font=huge}
        \caption{Inverted Pixels}
        \label{fig7b}
      \end{subfigure}
    }
   \resizebox{0.5\textwidth}{!}{
      \begin{subfigure}[b]{0.5\textwidth}
        \centering
        \includegraphics[width=\textwidth, height = 70pt]{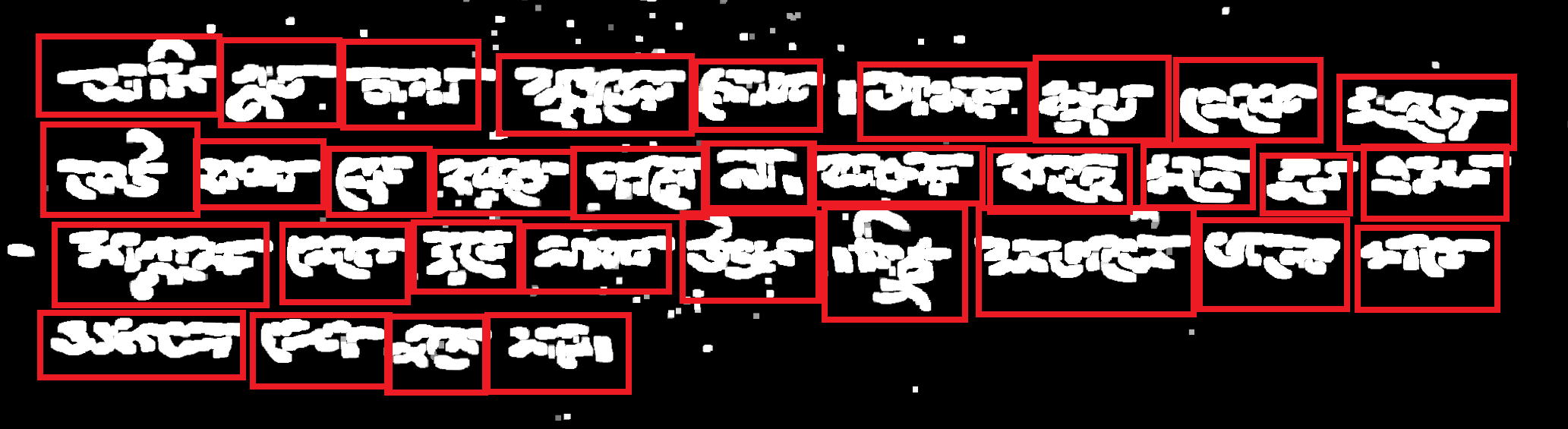}
        \captionsetup{font=huge}
        \caption{kernel size = 3 * 3 , iteration = 3, No. of words = 33}
        \label{fig7e}
      \end{subfigure}
      \begin{subfigure}[b]{0.5\textwidth}
        \centering
        \includegraphics[width=\textwidth, height = 70pt]{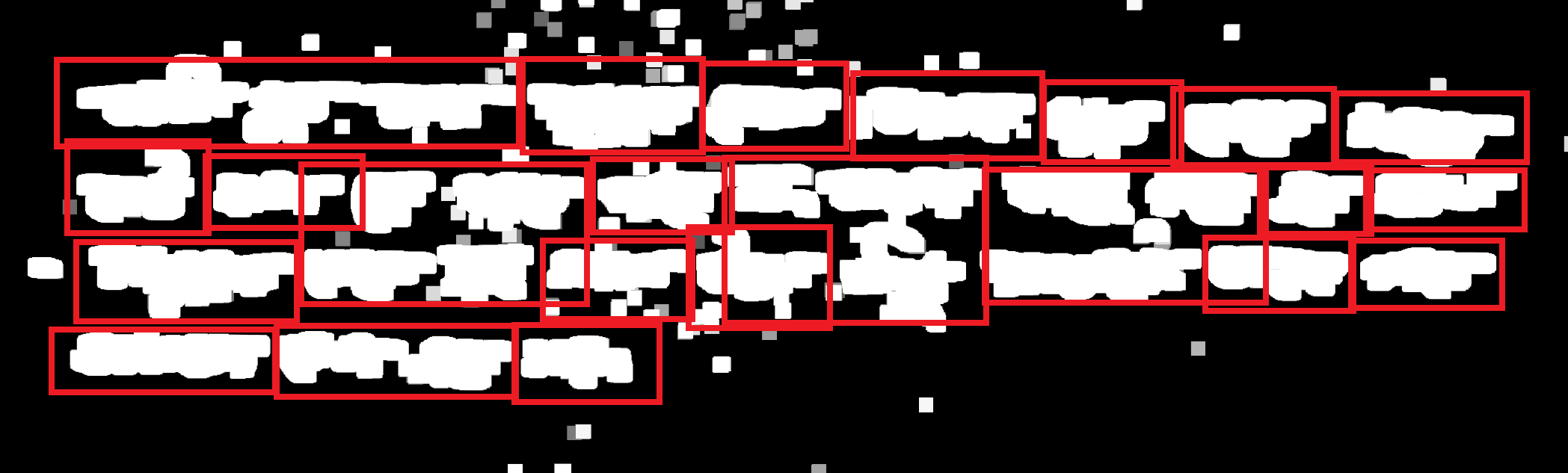}
        \captionsetup{font=huge}
        \caption{kernel size = 10 * 10 , iteration = 2, No. of words=23}
        \label{fig7h}
      \end{subfigure}
    }
  \caption{Result after applying dilation with different kernel size and iteration number}
  \label{fig7}
\end{figure}

\paragraph {Word Detection Using Contour Logic} After dilation, the words/contours in that image that are greater than the MAT are counted. Then, again dilation is applied on the same image with either more iteration or a larger kernel size, and the words are counted again.    
\paragraph{Finding Optimal Solution} 
This checking continues until the maximum number of words is found and stops just before two words collide during dilation. For the image in Fig. \ref{fig7b}, the ACS is 641. So, the starting kernel for dilation is 3*3. Most words (Fig. \ref{fig7e}) are detected when the optimal solution is reached. After this, the more iteration is added or kernel size is increased, the result constantly moves away from the optimal solution as shown in Fig. \ref{fig7h}.
\subsubsection{Word Segmentation Using Bounding Boxes}
\begin{figure}
  \centering
  \begin{subfigure}[b]{0.45\textwidth}
    \centering
    \includegraphics[width=0.9\textwidth, height = 30pt]{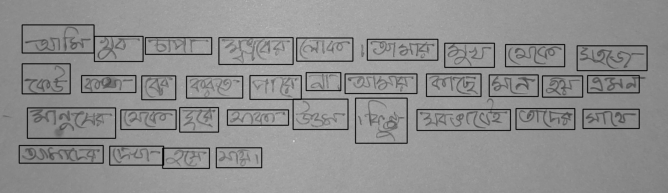}
  \end{subfigure}
  \caption{Words detected and surrounded by bounding boxes after applying all the stages on Fig. \ref{fig2a}}
  \label{fig8}
\end{figure}

Finally, when the Word Detection stage finds the optimal solution, Fig. \ref{fig8} shows all the words are segmented by bounding boxes after applying all the stages on the input image of Fig. \ref{fig2a}.

\begin{table}[ht]
  \centering
  \caption{Mapping of WD to MAT}
  \renewcommand{\arraystretch}{1.5}
  \begin{tabularx}{0.4819\textwidth}{|m{3cm}|m{4.875cm}|}
    \hline
    \centering Writing Density (WD) & \centering Minimum Area Threshold (MAT) \tabularnewline
    \hline
    \centering WD $<$ 1500 & \centering ACS - $\frac{ACS}{4}$ \tabularnewline
    \hline
    \centering $1500 \leq$ WD $<$ 2500 & \centering ACS * 2 - $\frac{ACS}{3}$ \tabularnewline
    \hline
    \centering $2500 \leq$ WD $<$ 4000 & \centering ACS * 3 - $\frac{ACS}{3}$ \tabularnewline
    \hline
    \centering $4000 \leq$ WD $<$ 6000 & \centering ACS * 4 - $\frac{ACS}{4}$ \tabularnewline
    \hline
    \centering $6000 \leq$ WD $<$ 8000 & \centering ACS * 4 \tabularnewline
    \hline
    \centering $8000 \leq$ WD $<$ 10000 & \centering ACS * 5 - $\frac{ACS}{2}$ \tabularnewline
    \hline
    \centering $10000 \leq$ WD & \centering ACS * 6 \tabularnewline
    \hline
  \end{tabularx}
  
  \label{table3}
\end{table}

\begin{table}[ht]
  \centering
  \caption{Further mapping of ACS and WD to MAT}
  \renewcommand{\arraystretch}{1.5}
  
  \begin{tabularx}{0.5\textwidth}{|m{1.71cm}|m{2.60cm}|m{3.49cm}|}
    \hline
    \centering Writing Density(WD) & \centering Average Contour Size (ACS) & \centering Minimum Area Threshold (MAT) \tabularnewline
    \hline
    \centering $2000 < $ WD & \centering $700 < $ ACS & \centering MAT = MAT - $\frac{ACS}{8}$ \tabularnewline
    \hline
    \centering $7000 < $ WD & \centering $1100 < $ ACS $\leq 1500$ & \centering MAT = MAT + ACS \tabularnewline
    \hline
    \centering $7000 < $ WD & \centering $1500 < $ ACS & \centering MAT = MAT + ACS * 1.5 \tabularnewline
    \hline
  \end{tabularx}
  
  \label{table4}
\end{table}
\begin{table}[ht]
  \centering
  \caption{Mapping of ACS to Starting Dilation Kernel Size}
  \begin{tabularx}{0.5\textwidth}{|m{3cm}|m{2.23cm}|m{2.5cm}|}
    \hline
    \centering Average Contour Size (ACS) & \centering Starting Dilation Kernel Size (Otsu) & \centering Starting Dilation Kernel Size (Adaptive Thresholding) \tabularnewline
    \hline
    \centering ACS $<$ 100 & \centering 1 $\times$ 1 & \centering 1 $\times$ 1 \tabularnewline
    \hline
    \centering $100 \leq$ ACS $<$ 1000 & \centering 3 $\times$ 3 & \centering 3 $\times$ 3 \tabularnewline
    \hline
    \centering $1000 \leq$ ACS $<$ 1300 & \centering 5 $\times$ 5 & \centering 5 $\times$ 5 \tabularnewline
    \hline
    \centering $1300 \leq$ ACS $<$ 1500 & \centering 7 $\times$ 7 & \centering 7 $\times$ 7 \tabularnewline
    \hline
    \centering $1500 \leq$ ACS $<$ 1800 & \centering 9 $\times$ 9 & \centering 9 $\times$ 9 \tabularnewline
    \hline
    \centering $1800 \leq$ ACS $<$ 1900 & \centering 10 $\times$ 10 & \centering 10 $\times$ 10 \tabularnewline
    \hline
    \centering $1900 \leq$ ACS $<$ 2000 & \centering 11 $\times$ 11 & \centering 11 $\times$ 11 \tabularnewline
    \hline
    \centering $2000 \leq$ ACS $<$ 2200 & \centering 9 $\times$ 9 & \centering 11 $\times$ 11 \tabularnewline
    \hline
    \centering $2200 \leq$ ACS $<$ 2400 & \centering 10 $\times$ 10 & \centering 12 $\times$ 12 \tabularnewline
    \hline
    \centering $2400 \leq$ ACS $<$ 2600 & \centering 11 $\times$ 11 & \centering 13 $\times$ 13 \tabularnewline
    \hline
  \end{tabularx}
  
  \label{table2}
\end{table}

\section{Results}
\subsection{Datasets}
\begin{figure}
  \centering
  \resizebox{0.5\textwidth}{!}{
    \begin{subfigure}[b]{0.2\textwidth}
      \centering
      \includegraphics[width=0.9\textwidth,height=120pt]{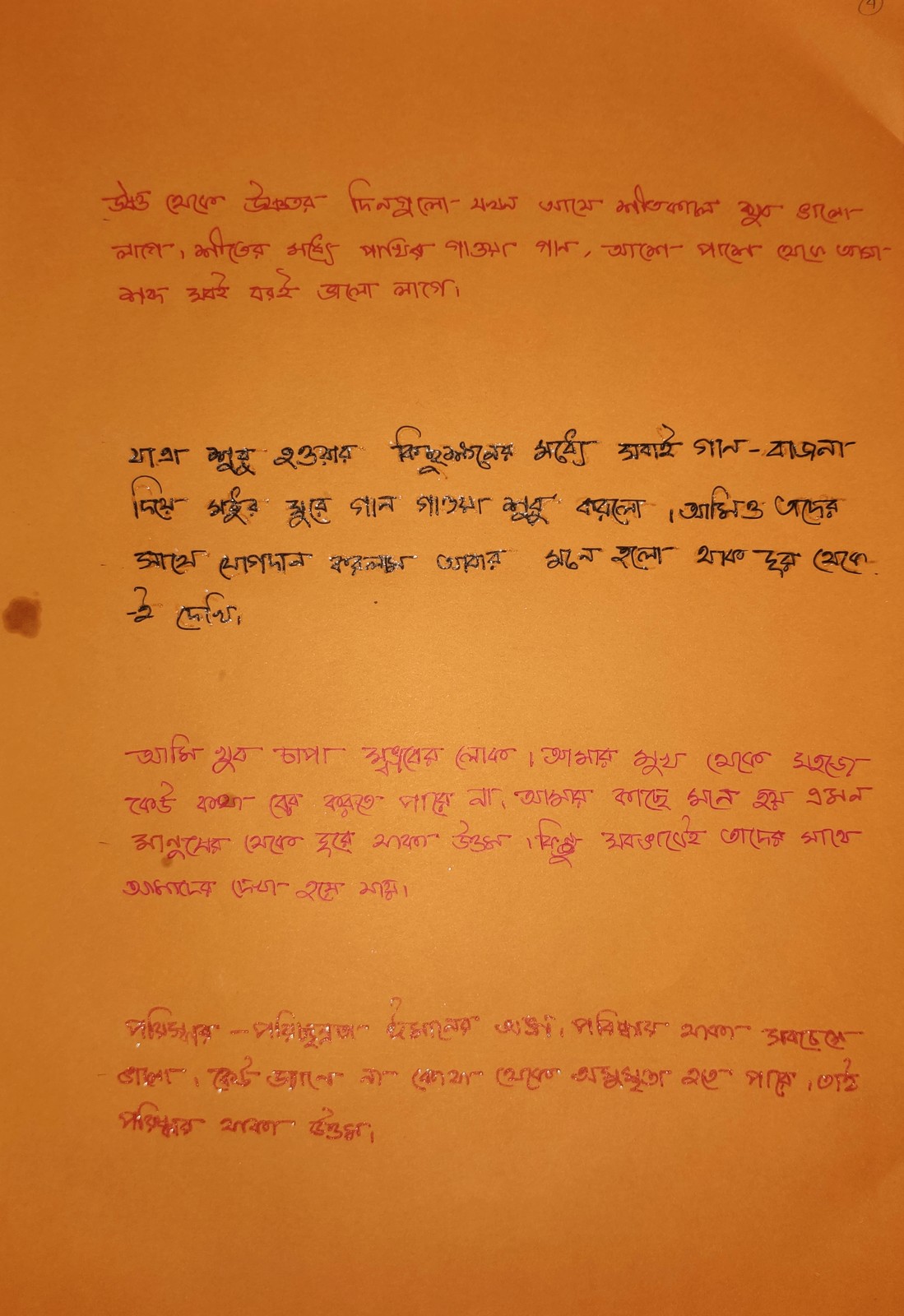}
    \label{fig9a}
    \end{subfigure}
    \hfill
    \begin{subfigure}[b]{0.2\textwidth}
      \centering
      \includegraphics[width=0.9\textwidth,height=120pt]{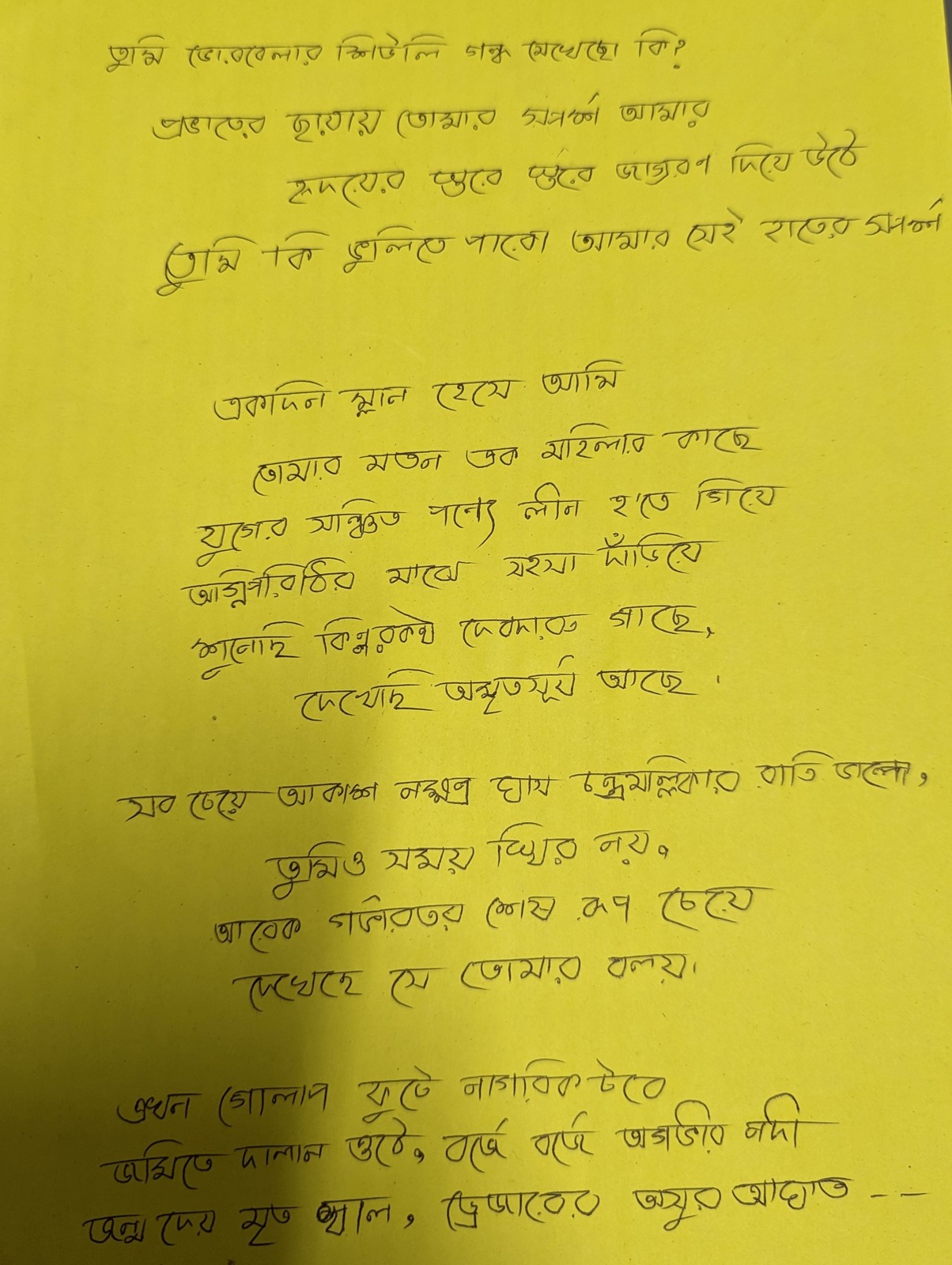}
    \label{fig9b}
      
    \end{subfigure}
    \begin{subfigure}[b]{0.2\textwidth}
      \centering
      \includegraphics[width=0.9\textwidth,height=120pt]{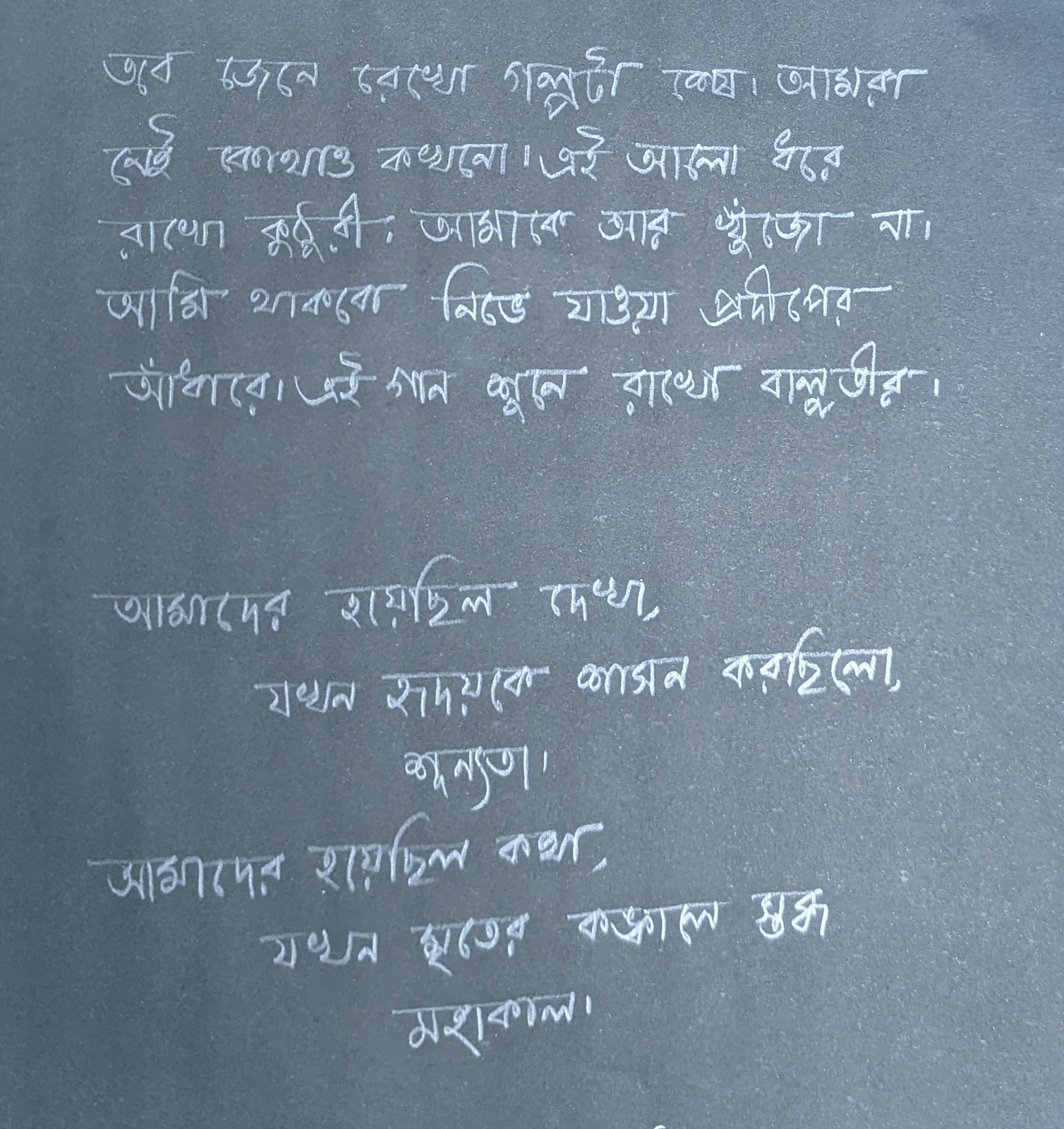}
    \label{fig9b}
      
    \end{subfigure}
    \begin{subfigure}[b]{0.2\textwidth}
      \centering
      \includegraphics[width=0.9\textwidth,height=120pt]{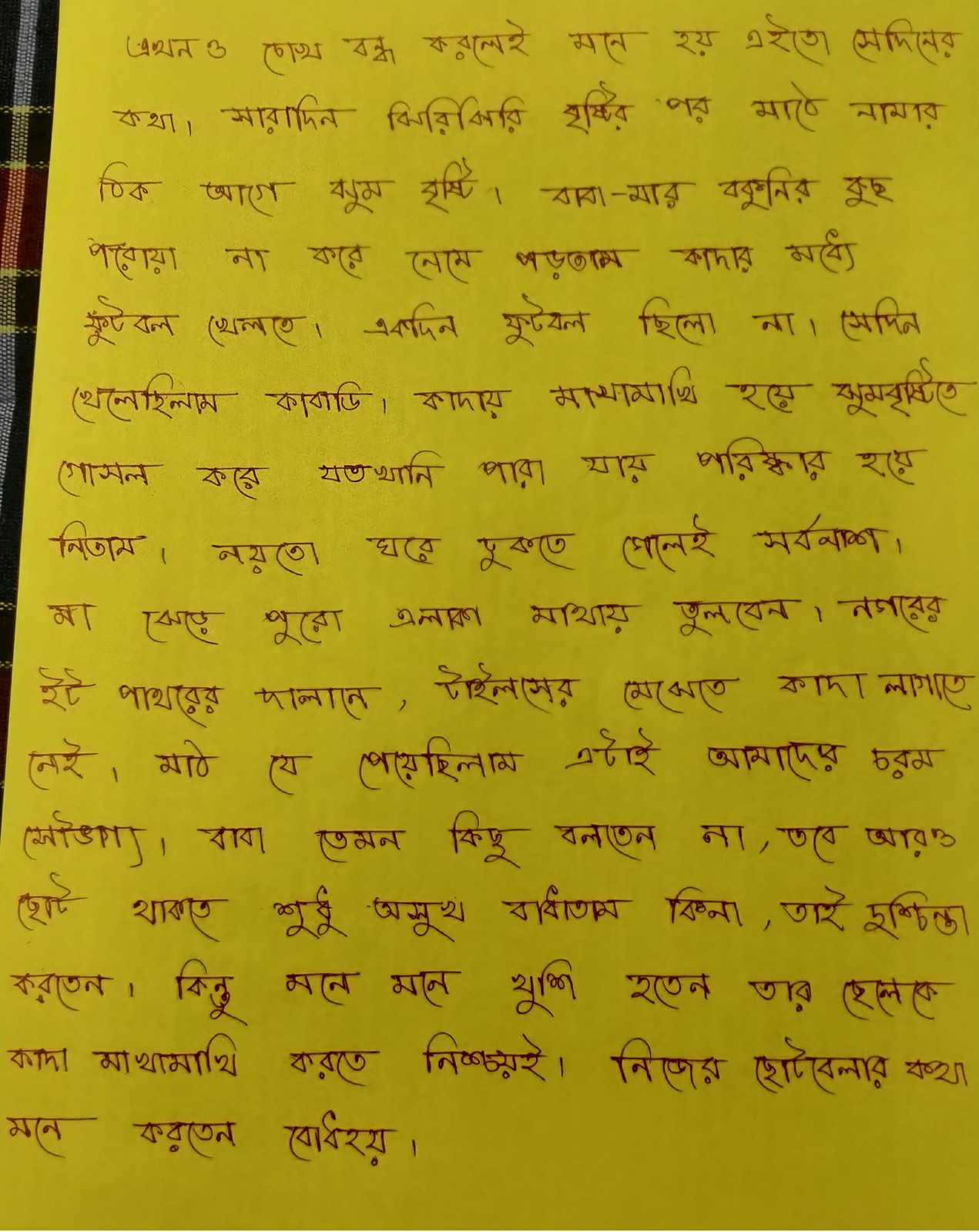}
    \label{fig9b}
      
    \end{subfigure}
  }
  \caption{Sample Images from the Customized Dataset}
  \label{fig9}
\end{figure}

 The authors created a customized dataset of 80 handwritten Bangla documents with various handwriting styles for the Word Segmentation stage. The papers used here are of different colors, including white, red, blue, green, and black. Different colors of glittering, gel and ballpoint pens were also used, such as black, green, red, blue, and green, to create a variation and build a generalized word detection system. Also, shadow interference is introduced as it is normal to have illumination issues in a smartphone-captured picture. Different kinds of devices are used so that each picture has a different size and resolution.  Fig. \ref{fig9} shows some samples of the customized Dataset.

\subsection{Evaluation Metrics}
Various metrics such as Recall, Precision, and F1-score are used to measure the accuracy of the system's detection rate. 
Recall\eqref{eq1} is calculated by dividing the number of right results by the number of results that should be returned. Precision\eqref{eq2} is the number of correct results divided by the total number of results. F-measure (F1 score)\eqref{eq3} is a weighted harmonic mean of precision and recall.
\subsection{Experimental Results of Word Segmentation}
In this experiment, 80 Bangla document images consisting of 7374 words are created to detect words using the abovementioned processes. A total of 7278 bounding boxes are generated. Of 7374 words, 6681 are correctly segmented by bounding boxes, 527 are either under-segmented, over-segmented or defect-segmented, and 166 are undetected. Recall, Precision, and F1-score are measured and reported in TABLE \ref{table5}. The formulas used to calculate them are as follows : 
\begin{equation}
\text{Recall} = \frac{\text{Total no. of correctly segmented words}}{\text{Total no. of actual words present}} \label{eq1}
\end{equation}
\begin{equation}
\text{Precision} = \frac{\text{Total no. of correctly segmented words}}{\text{Total no. of bounding boxes generated}} \label{eq2}
\end{equation}
\begin{equation}
\text{F1-score} = 2 \times \frac{\text{Precision} \times \text{Recall}}{\text{Precision} + \text{Recall}} \label{eq3}
\end{equation}

  

\begin{table}[b]
  \centering
  \caption{Comparative Evaluation: Recall, Precision, and F1-Score of the proposed method across various datasets and comparison with existing approaches}
  \renewcommand{\arraystretch}{1.5}
  \begin{tabularx}{0.5\textwidth}{|m{1.50cm}|m{0.82cm}|m{1.0cm}|m{0.82cm}|m{1.20cm}|m{1.10cm}|}
    \hline
    \centering Methodology & \multicolumn{3}{|c|}{Proposed} & \centering Sarkar et al.\cite{id12} & \centering Das et al.\cite{id15} \tabularnewline
    \hline
 
    \centering Evaluation Metric & \centering Recall (\%) & \centering Precision (\%) & \centering F1 score (\%) & \centering Success-Rate (\%) & \centering Average Accuracy (\%) \tabularnewline
    \hline
    \centering Created dataset & \centering $90.60$ & \centering $91.80$ & \centering $91.20$ & & \tabularnewline
    \hline
    \centering CMATERdb 1.1.1\cite{id17} dataset & \centering $86.15$ & \centering $86.94$ & \centering $86.54$ & \centering $86.01$ & \tabularnewline
    \hline
    \centering Bangla Writing dataset\cite{id15} & \centering $86.22$ & \centering $88.72$ & \centering $87.45$ & & \centering $84.55$ \tabularnewline
    \hline
  \end{tabularx}
  
  \label{table5}
\end{table}

The proposed methodology is also applied to 100 documents from the CMATERdb1.1.1\cite{id17} dataset, and a total of 15587 bounding boxes are generated. Of 15730 words, 13552 are correctly segmented by bounding boxes, 2048 are either under-segmented, over-segmented or defect-segmented, and 130 are undetected. Sarkar et al.\cite{id12} also worked on the same dataset, but their input was document images which had already been segmented into text lines. Out of 15730 words, they correctly detected 13530, and 2200 words were either under-segmented or over-segmented. Despite working with segmented lines, They had a success-rate of 86.01 \% for word segmentation. This result would degrade even more if they worked with paragraph images, rather than images of segmented text lines. TABLE \ref{table5} shows a result comparison between the proposed methodology and Sarkar et al.\cite{id12}'s work.

The proposed methodology is also applied to 100 documents from the Bangla Writing dataset\cite{id15}, and a total of 7191 bounding boxes are generated. Of 7400 words, 6380 are correctly segmented by bounding boxes, 987 are either under-segmented, over-segmented or defect-segmented, and 33 are undetected. Das et al.\cite{id15} worked on a portion of this dataset, finding a line segmentation accuracy of 88.39\% , word segmentation accuracy of 95.65\% , and average accuracy of 84.55\%. TABLE \ref{table5} shows the result comprarison between the proposed methodology and Das et al.\cite{id15}'s approach.

\begin{figure}
  \centering
  \resizebox{0.5\textwidth}{!}{
      \begin{subfigure}[b]{0.5\textwidth}
        \centering
        \includegraphics[width=\textwidth, height = 30pt]{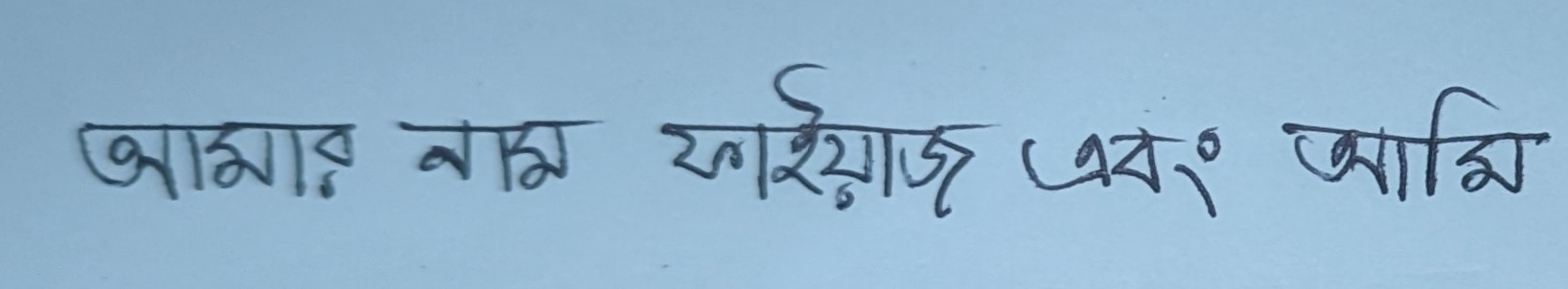}
        \captionsetup{font=huge}
        \caption{Input image}
        \label{fig10a}
      \end{subfigure}
      \begin{subfigure}[b]{0.5\textwidth}
        \centering
        \includegraphics[width=\textwidth, height = 30pt]{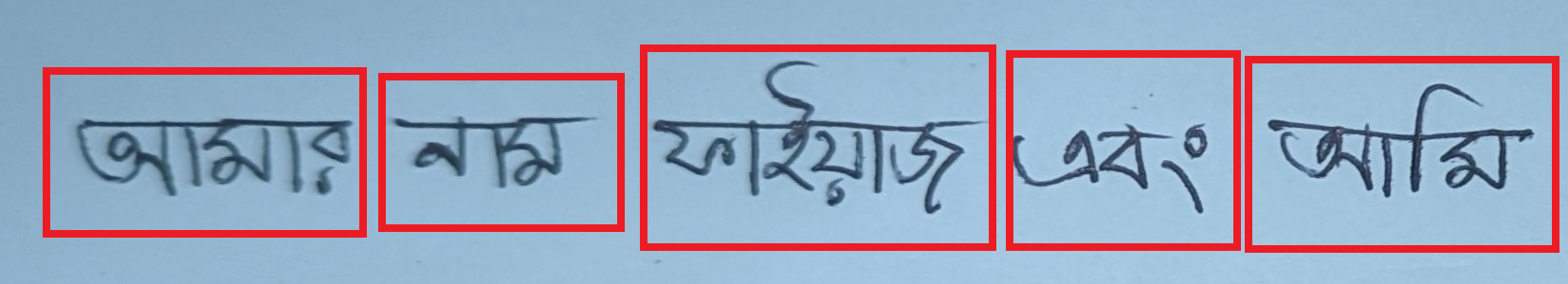}
        \captionsetup{font=huge}
        \caption{No error}
        \label{fig10b}
      \end{subfigure}
      \begin{subfigure}[b]{0.5\textwidth}
        \centering
        \includegraphics[width=\textwidth, height = 30pt]{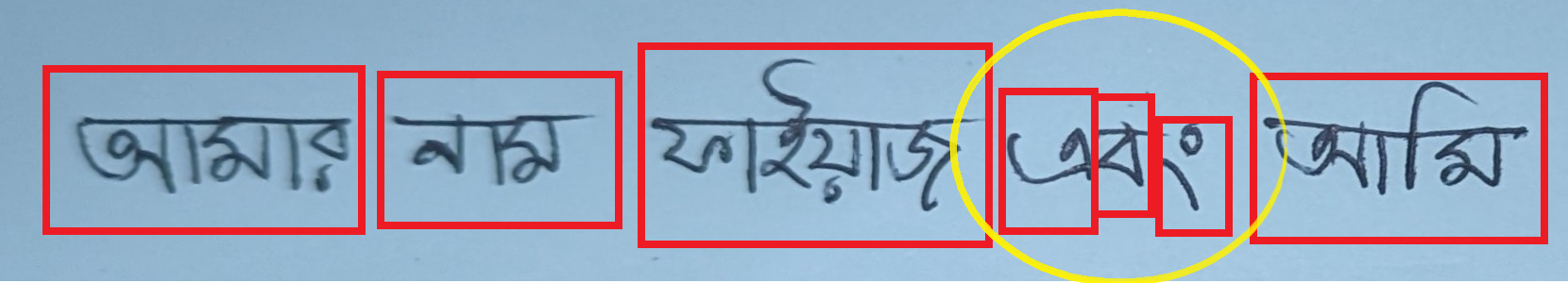}
        \captionsetup{font=huge}
        \caption{Under-segmentation error}
        \label{fig10c}
      \end{subfigure}
}
  \resizebox{0.5\textwidth}{!}{
      \begin{subfigure}[b]{0.5\textwidth}
        \centering
        \includegraphics[width=\textwidth, height = 30pt]{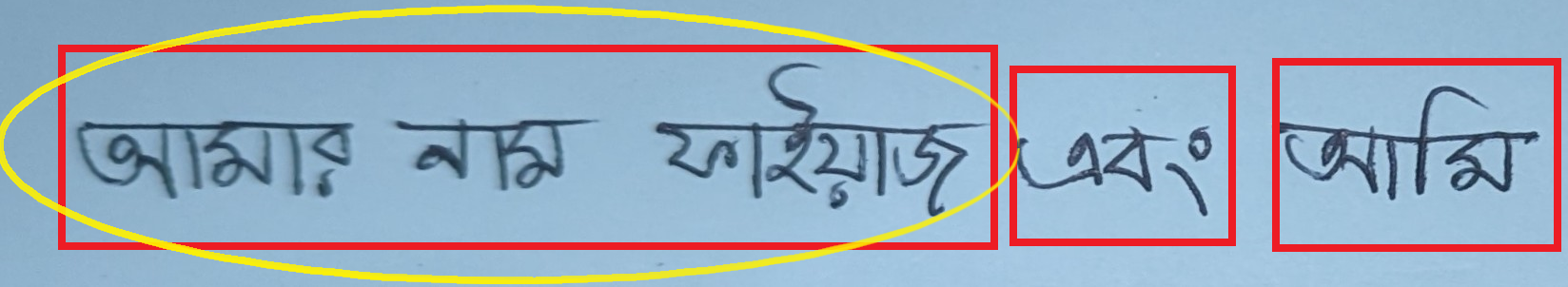}
        \captionsetup{font=huge}
        \caption{Over-segmentation error}
        \label{fig10d}
      \end{subfigure}
      \begin{subfigure}[b]{0.5\textwidth}
        \centering
        \includegraphics[width=\textwidth, height = 30pt]{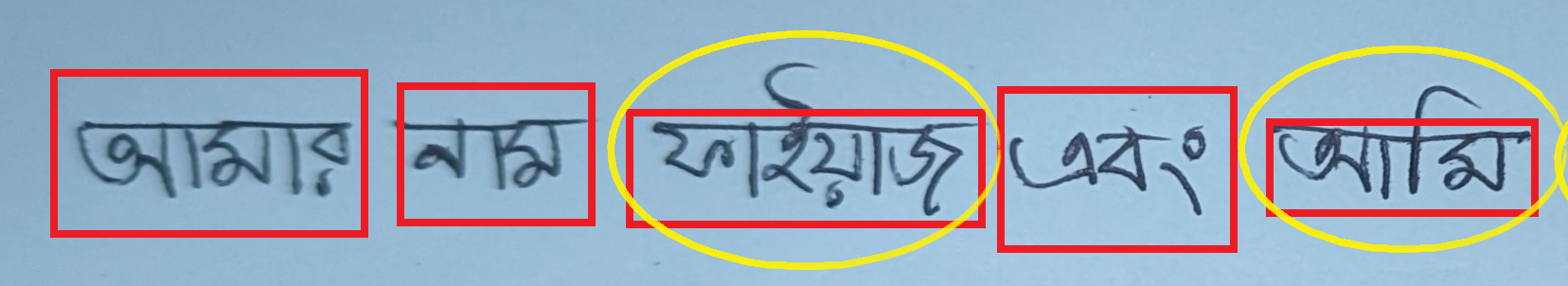}
        \captionsetup{font=huge}
        \caption{Defect-segmentation error}
        \label{fig10e}
      \end{subfigure}
      \begin{subfigure}[b]{0.5\textwidth}
        \centering
        \includegraphics[width=\textwidth, height = 30pt]{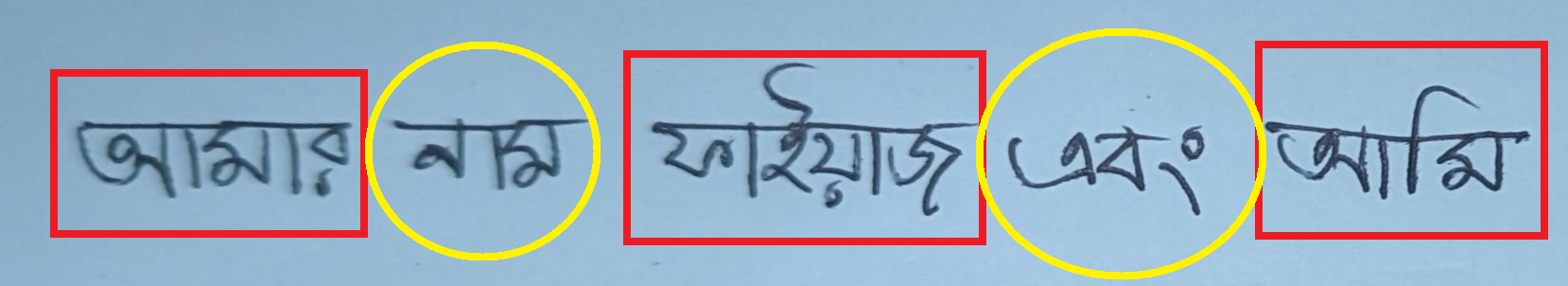}
        \captionsetup{font=huge}
        \caption{Overlooking error}
        \label{fig10f}
      \end{subfigure}
  }
  \caption{Demonstration of different kinds of errors encountered during the proposed word detection approach (errors are highlighted by yellow markers)}
  \label{fig10}
\end{figure}

Most words are accurately identified, but there are some undetected words and instances where a single bounding box is drawn across multiple words, or multiple bounding boxes detect a single word. To analyze the problems, four kinds of errors can be introduced,e.g., under-segmentation error, over-segmentation error, defect-segmentation error, and overlooking error. Under-segmentation error means a single word is erroneously broken down into two/more parts and bounded by two/more bounding boxes. An over-segmentation error means a single bounding box detects two/more words. Defect-segmentation error means a single bounding box bounds a word, but some part of it,e.g., \textbf{\bengalifont'া', 'ি,' 'ঁ,' 'ৃ,' 'ী'} (primarily vowels) is excluded. Fig. \ref{fig10} shows the demonstration of these errors. Besides, punctuation marks, e.g., \textbf{\bengalifont '।', ','} etc., are detected along with their nearby words. But in some handwriting styles, they can be detected by an individual bounding box.

\begin{figure}
  \centering
  \resizebox{0.5\textwidth}{!}{
      \begin{subfigure}[b]{0.45\textwidth}
        \centering
        \includegraphics[width=0.85\textwidth, height= 20pt]{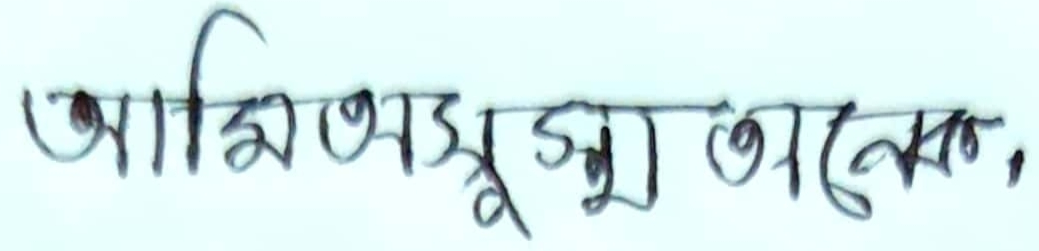}
        \captionsetup{font=Large}
        \caption{\Large Over-segmentation}
        \label{fig11a}
      \end{subfigure}
      \begin{subfigure}[b]{0.45\textwidth}
        \centering
        \includegraphics[width=0.85\textwidth, height= 20pt]{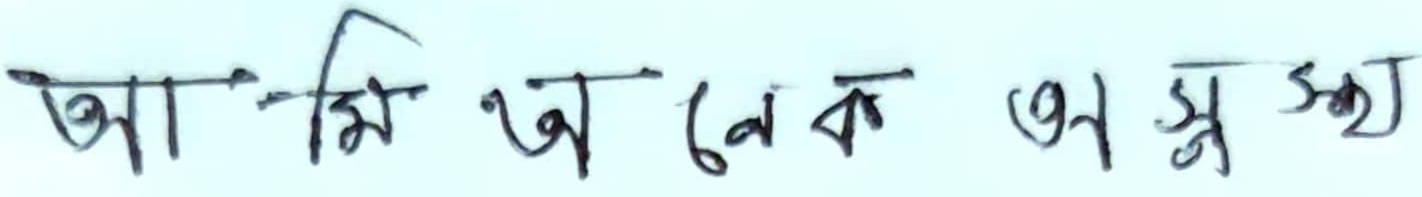}
        \captionsetup{font=Large}
        \caption{\Large Under-segmentation }
        \label{fig11b}
      \end{subfigure}
  }
  \resizebox{0.5\textwidth}{!}{
      \begin{subfigure}[b]{0.45\textwidth}
        \centering
        \includegraphics[width=0.85\textwidth, height= 20pt]{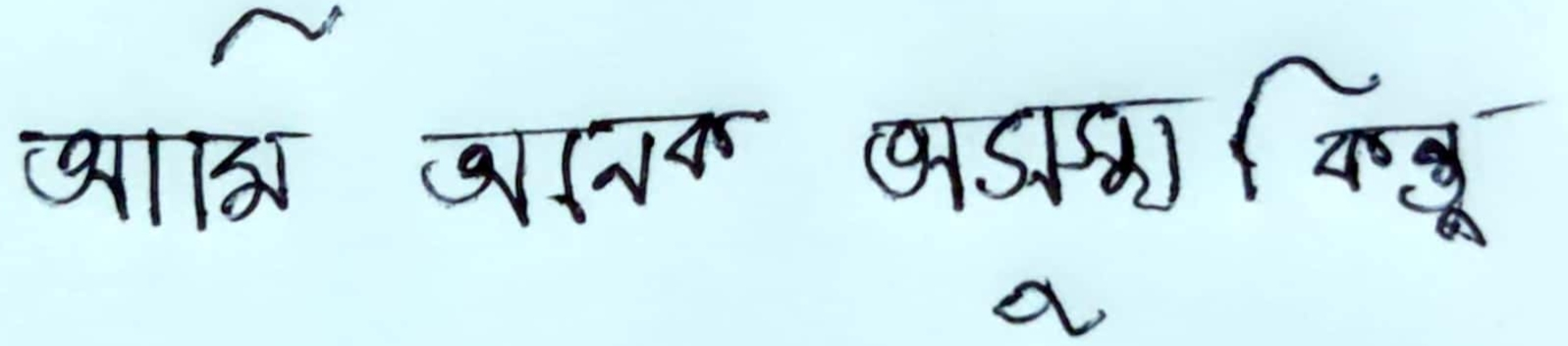}
        \captionsetup{font=Large}
        \caption{\Large Defect-segmentation }
        \label{fig11c}
      \end{subfigure}
      \begin{subfigure}[b]{0.45\textwidth}
        \centering
        \includegraphics[width=0.85\textwidth, height= 20pt]{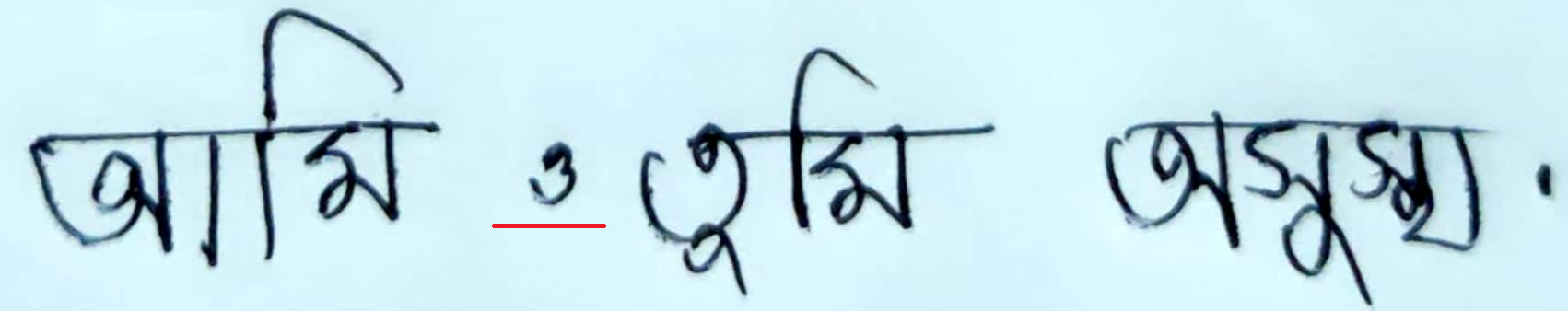}
        \captionsetup{font=Large}
        \caption{\Large Overlooking }
        \label{fig11d}
      \end{subfigure}
  }
  \caption{Demonstration of different kinds of handwritings that are prone to generate different errors}
  \label{fig11}
\end{figure}

Different people have different writing styles. Some people write so densely that there are no sufficient gaps between separate lines and words, e.g., Fig. \ref{fig11a}. It increases the chance of over-segmentation errors. In that case, if a smaller starting dilation kernel or smaller MAT is used at first, it can reduce that error. However, it increases the chance of under-segmentation error when another writer writes in a scattered fashion, so words have huge gaps between the letters, e.g., Fig. \ref{fig11b}. On the other hand, using a smaller dilation kernel can lead to more defect-segmentation errors as some people prefer to write in a fashion that the vowels, e.g., \textbf{\bengalifont'া,' 'ি,' 'ঁ,' 'ৃ,' 'ী'} are very distant from the actual word, e.g., Fig. \ref{fig11c}. So, the vowels are excluded from the bounding box. It can be improved by introducing a larger MAT. However, it increases the chance of over-segmentation error and overlooking error for words with smaller areas, e.g., Fig. \ref{fig11d}.
In this work, an optimal solution has been chosen so that no error goes to an extreme extent. However, if a writer writes in such a fashion that there are hardly any gaps between words/lines or a large gap inside a word, it will most likely generate errors. 

\section{Conclusion}
OCR systems use computer vision to recognize and transform text images into editable text. In this research, a word segmentation system for smartphone-captured images of handwritten Bangla documents is created, which uses a bespoke dataset and some image processing techniques to segment words from paragraphs. However, the system has limitations too, e.g., if there is so little gap between words and lines that it is impossible even for a human to detect separate entities, it can lead to over-segmentation error. It can be solved by introducing a machine-learning-based model where the model is trained with every word available in the dictionary. If a person's writing style is such that he does not use matra at all or the characters are pretty distant from one another, it can lead to under-segmentation errors. Some Bangla characters, e.g.,\textbf{\bengalifont'ঃ ,' 'ং,' 'ঁ'}, do not have matra. These characters can sometimes lead to under-segmentation errors even while maintaining a good handwriting style. If there is some background behind the paper which has a pixel value almost similar to the characters' pixel value, the system detects those background areas as words. It can be solved by cropping the picture to eliminate the background.

The proposed method can perform best if the writer correctly maintains the inter-word/line/character gaps. However, the most promising fact about this work is that it can take smartphone-captured images written on any color of paper, with any color of ink. Also, another fact is that an over-segmentation error does not ruin two lines because it does not deal with any line segmentation method. Instead, it creates an over-segmentation error between two/more words. Also, a further nested approach can be introduced to deal with the over-segmentation errors, increasing the system's accuracy even more.
The system's output has been compared with some existing works and performs better than those. The segmented words can be further used to segment characters or to detect words from them. The current system could aid in the transition to a more established OCR system.

\bibliographystyle{IEEEtran}
\bibliography{references}

\end{document}